%% file: ms.tex
\documentclass[preprint]{vgtc}               

\ifpdf
  \pdfoutput=1\relax                   
  \usepackage{graphicx}                
  \DeclareGraphicsExtensions{.pdf,.png,.jpg,.jpeg} 
\else
  \usepackage{graphicx}                
  \DeclareGraphicsExtensions{.eps}     
\fi%

\graphicspath{{figs/}{pictures/}{images/}{./}} 

\usepackage{microtype}                 
\PassOptionsToPackage{warn}{textcomp}  
\usepackage{textcomp}                  
\usepackage{mathptmx}                  
\usepackage{times}                     
\usepackage{cite}                      
\usepackage{tabu}                      
\usepackage{booktabs}                  

\usepackage{amsmath}
\usepackage{balance}  
\usepackage{txfonts}
\usepackage[usenames,dvipsnames]{xcolor}
\usepackage{color}
\usepackage{xspace}
\usepackage{csquotes}

\newcommand{\comment}[2]{#2}

\newcommand{\joschi}[1]{\comment{NavyBlue}{#1}}
\newcommand{\enrico}[1]{\comment{Orange}{#1}}
\newcommand{\aritra}[1]{\comment{Purple}{#1}}
\newcommand{\aritranew}[1]{\comment{Red}{#1}}

\newcommand{\eg}{\emph{e.g.}\xspace}
\newcommand{\etal}{\emph{et al.}\xspace}
\newcommand{\ie}{\emph{i.e.}\xspace}

\newcommand{\tabA}{Statistical Summary View\xspace}
\newcommand{\tabB}{Explanation Explorer\xspace}
\newcommand{\tabC}{Item Level Inspector\xspace}

\onlineid{0}

\vgtccategory{Research}

\vgtcinsertpkg

\usepackage{caption,subcaption}

\preprinttext{Published at the IEEE Conference on Visual Analytics Science and Technology (IEEE VAST 2017).}

\title{A~Workflow~for~Visual~Diagnostics~of~Binary~Classifiers using~Instance-Level~Explanations}

\author{
Josua Krause\thanks{e-mail: josua.krause@nyu.edu}\\ %
\parbox{1.0in}{\scriptsize \centering NYU~Tandon School~of~Engineering} %
\and Aritra Dasgupta\thanks{e-mail: aritra.dasgupta@pnnl.gov}\\ %
\parbox{1.0in}{\scriptsize \centering Pacific~Northwest National~Laboratory} %
\and Jordan Swartz\thanks{e-mail: jordan.swartz@nyumc.org}\\ %
\parbox{1.0in}{\scriptsize \centering NYU School~of~Medicine} %
\and Yindalon Aphinyanaphongs\thanks{e-mail: yin.a@nyumc.org}\\ %
\parbox{1.0in}{\scriptsize \centering NYU School~of~Medicine} %
\and Enrico Bertini\thanks{e-mail: enrico.bertini@nyu.edu}\\ %
\parbox{1.0in}{\scriptsize \centering NYU~Tandon School~of~Engineering} %
}

\input{tex/abstract} 


\begin{document}

\firstsection{Introduction}

\maketitle


\input{tex/introduction}

\input{tex/relwork}

\input{tex/motivation}

\input{tex/algo}
\input{tex/ui}
\input{tex/case}

\input{tex/conclusion}

\acknowledgments{
\aritra{We thank Prof. Foster Provost for his help in understanding and using his instance-level explanation technique. The research described in this paper is part of the Analysis in Motion Initiative at Pacific Northwest National Laboratory (PNNL). It was conducted under the Laboratory Directed Research and Development Program at PNNL, a multi-program national laboratory operated by Battelle. Battelle operates PNNL for the U.S. Department of Energy (DOE) under contract DE-AC05-76RLO01830. The work has also been partially funded by the Google Faculty Research Award "Interactive Visual Explanation of Classification Models".}}

\bibliographystyle{abbrv-doi}

\balance
\bibliography{ms}
\end{document}

%% file: tex/abstract.tex
\abstract{Human-in-the-loop data analysis applications necessitate greater transparency in machine learning models for experts to understand and trust their decisions. To this end,   
we propose a visual analytics workflow to help data scientists and domain experts explore, diagnose, and understand the decisions made by a binary classifier.
The approach leverages ``instance-level explanations", measures of local feature relevance that explain single instances, and uses them to build a set of visual representations that guide the users in their investigation.
The workflow is based on three main visual representations and steps:
one based on aggregate statistics to see how data distributes across correct / incorrect decisions;
one based on explanations to understand which features are used to make these decisions;
and one based on raw data, to derive insights on potential root causes for the observed patterns.
The workflow is \aritra{derived from} a long-term collaboration with a group of machine learning and healthcare professionals who used our method to make sense of machine learning models they developed.
\joschi{The case study from this collaboration demonstrates that the proposed workflow helps experts derive useful knowledge about the model and the phenomena it describes, thus experts can generate useful hypotheses on how a model can be improved.}
}

%% file: tex/introduction.tex
In this paper we propose an interactive workflow and a visual user interface to help data scientists and domain experts diagnose and validate binary classifiers. The approach we suggest is based on a mix of automated and interactive methods that guide the user towards understanding what decisions a model makes, which ones are correct or incorrect, and potential strategies to improve them.

Being able to explore the decisions a model makes and identifying potential issues is crucial in application areas where experts need to get a sense of how the model works and build trust in its decisions. While common practice in much of the machine learning endeavors is to focus on model accuracy, many researchers have voiced the need for more transparency when the application domain requires it~\cite{baesens2003using, Caruana:2015:IMH:2783258.2788613, freitas2014comprehensible, lou2012intelligible, martens2007comprehensible, vellido2012making}. A recent DARPA (Defense Advanced Research Projects Agency) program called ``Explainable AI (XAI)", for example, calls for more research in this area and declares, as the main motivation for the program that ``\textit{the effectiveness of these systems is limited by the machine’s current inability to explain their decisions and actions to human users}" and that ``\textit{it is essential to understand, appropriately trust, and effectively manage an emerging generation of artificially intelligent machine partners}".

In addition to evaluating a model in terms of accuracy, we propose the idea of \textit{semantic validation}, the need for domain experts to verify that the decisions a model makes are plausible when compared against their mental models of the problem. For instance, in healthcare settings, medical doctors often want to see examples of recommendations the model provides and need to gain trust in it before they feel comfortable with deploying it in real-world settings.
Such reservations in deploying models without having an opportunity to manually verify what decisions they make are well justified as it is entirely possible for a model to achieve high accuracy and yet provide dramatically erroneous recommendations~\cite{Caruana:2015:IMH:2783258.2788613}.

Another important factor to consider is that domain experts and data scientists are often working in collaboration to solve a particular problem (or they are actually the same person covering both roles). Being able to manually inspect a model can give them an opportunity to generate useful insights on how a model can be improved. While commonly used aggregate statistics such as area under the curve (AUC) give a sense of the overall accuracy of the model, and can be used as a parameter to compare between different models, they do not provide insights on how or why a model fails to capture important phenomena accurately.

Some existing methods do provide more transparency and useful information for enabling better understanding and diagnostic purposes, but they tend to be limited and specific to a particular kind of model. For example, \textit{logistic regression} and \textit{decisions trees} are commonly regarded as more interpretable models thanks to their ability to provide information on feature weights and / or specific decisions the model makes (decision trees)~\cite{freitas2014comprehensible}. These solutions are however limited by a number of factors. Since they are specific to the selected method, they are hard to generalize and cannot be applied transparently to other types of models. Furthermore, they only provide a limited picture of what decisions the model makes. Feature weights provide a highly coarse summary of how relevant features are \textit{globally}, but they do not provide information on how the model makes decisions \textit{locally}, for a selected set of instances. Even more transparent methods, like decision trees, tend to grow very large and are not easy to parse visually, especially for data sets with a high number of dimensions / features.

\input{fig/workflow}

To address these issues we propose a workflow, \joschi{aimed at machine learning experts and data scientists}, based on \textit{instance-level explanations}, computational methods to derive a description of how a model makes decisions on single data items, without having access to the internal logic of the model (\ie, using the model as a \textit{black box}). These explanations are then aggregated and used as input to a visualization system that enables the browsing of model decisions and assessment of their quality.

The work we describe in the paper stems from a one year collaboration with a group of domain and machine learning experts~from the \textit{NYU Langone Medical Center}.
In our collaboration, we worked together to make sense of models built to understand how patients are handled in the hospital and to figure out whether important outcomes of interest can be predicted correctly. This resulted in the development of an interactive model diagnostic workflow using visual explanations of model behavior that is the main contribution of this work.
The rest of the paper is organized as follows. We present related work in the next section. We provide an overview of model diagnostics goals and of the proposed workflow in Section~\ref{sec:model-diagnostics}. We then describe in detail the instance-level explanation algorithm we use in Section~\ref{sec:algo} and the interfaces we built in Section~\ref{sec:ui}. Section~\ref{sec:case_study} reports on a use case we built to show how the workflow can help perform useful and actionable model diagnostics. Section~\ref{sec:discussion} discusses the results and provides a number of reflections and lessons we have learned from this collaborative exercise.

%% file: fig/workflow.tex
\begin{figure*}[t]
\centering
\includegraphics[width=\textwidth]{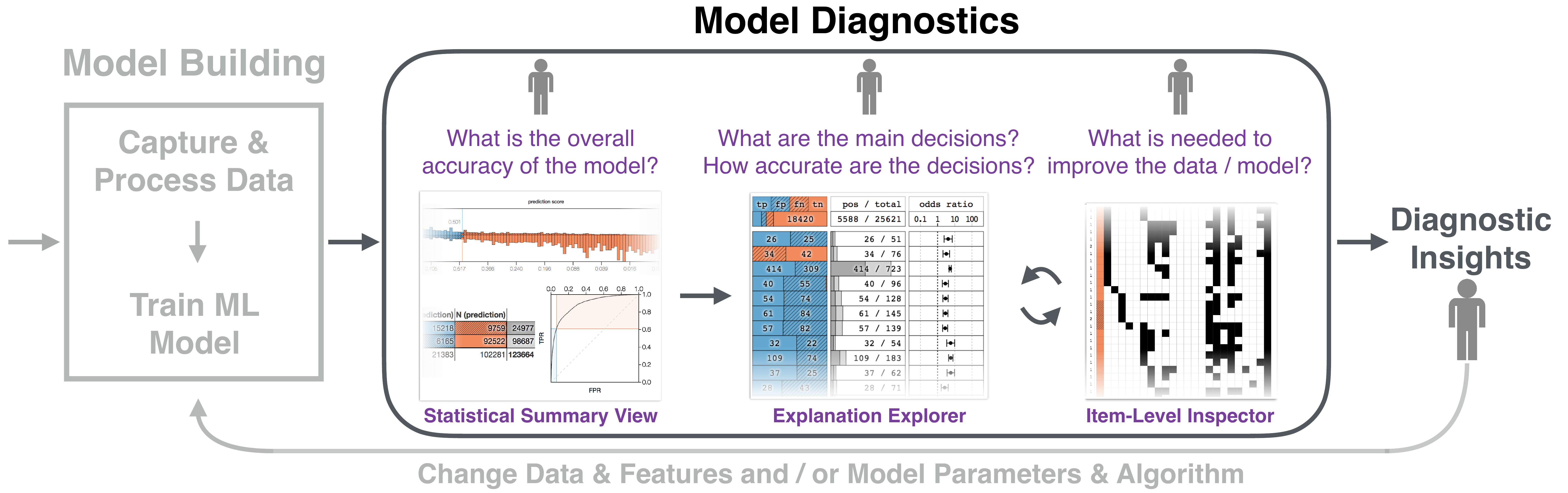}
\vspace{-3mm}
\caption{
Our proposed \textbf{Model Diagnostics} workflow extends the conventional \textit{Model Building} workflow in machine learning for enabling domain experts to reason about the semantic validity of the decisions made by any model through multiple linked visualizations of statistical performance summaries, explanations, and item-level distribution of features. By iterating through explanation-level summaries and item-level details, experts are able to generate diagnostic insights about the quality of both the data and the model. This ultimately helps to improve data acquisition and model generation processes belonging to the original workflow.
}
\vspace{-5mm}
\label{figs:workflow}
\end{figure*}


%% file: tex/relwork.tex
\section{Related Work}

In the following, we discuss model explanations and visual analytics techniques used for interacting with classification models.

\subsection{Model Explanations: \textit{Why} and \textit{How}}

Explanations of behavior of autonomous systems~\cite{Lim:2009:ADI:1620545.1620576} or computational models~\cite{dasgupta2017familiarity} can lead to a high degree of human-machine trust. In machine learning, model explanations are beginning to be used in human-in-the-loop data analysis applications for communicating information about model behavior and predictions. While there are some studies~\cite{harmful} that show that explanations can lead to over-reliance on the system, generally it has been posited that model explanations lead to a high degree of human interpretability and trust~\cite{lipton2016mythos}. Similar to the latter, our goal in this work was to develop a visual analytic workflow for model explanations and to work closely with data scientists and domain experts to understand how that could lead them to understand and trust model behavior.

In the literature, we find two contrasting purposes behind generating model explanations. The first approach is embedded within the interactive machine learning pipeline and helps end users in refining a model's predictions by interacting with the model structure. This helps users to build a mental model about the model reasoning process~\cite{Kulesza:2015:PED:2678025.2701399}. Through the EluciDebug approach, Kulesza \etal lay out a set of principles for the process of explanatory debugging using a Na\"ive Bayes classifier model.
Although the principles are generally applicable, the explanation technique is specifically applicable only to a particular model.
\joschi{Furthermore, additive models enable intuitive explanations through feature contributions that allow both to visualize the decision making process for single instances \cite{Poulin:2006:VEE:1597122.1597143} and feature contributions on a population level \cite{Caruana:2015:IMH:2783258.2788613}.
However, this solution requires the use of an additive model and as such it is not generally applicable.
}

To overcome this limitation, a second approach for explanation generation is to treat the machine learning model as a black box, bypassing the model structure, while communicating the input-output relationships and their relevance to a model's decisions to an analyst, \eg, inferring rules from a neural network~\cite{Craven98usingneural}, or generating explanations~\cite{DBLP:journals/corr/RibeiroSG16,prospector}. We adopt this black-box approach in our workflow for benefiting domain experts, who are not trained in machine learning, and also for providing data scientists with a model-agnostic and generalizable diagnostic interface for inspecting model quality. In previous work, local explanations have been used to diagnose how models make decisions for single instances of a data set~\cite{infuse,prospector,DBLP:journals/corr/RibeiroSG16}.
In contrast, we provide an interactive workflow where users can explore aggregated representations of explanations and better understand the context of model decisions by iterating across explanation-level and instance-level visual summaries of prediction quality.  




\subsection{Human-in-the-Loop Inspection of \joschi{Classifiers}}

Researchers have recently demonstrated how human interventions can help in greater accuracy in construction of classifiers, when compared with a purely automated approach~\cite{tam2017analysis}. In this work, Tam \etal used information theory to show how soft knowledge of model developers can be encoded in decision trees, and they advocate a tighter integration between human and machine-centric processes for model development. The goals for integrating visual analytic techniques and classification methods fall broadly into three categories, as proposed by Liu \etal~\cite{liu2017towards}: i) model understanding, ii) model diagnosis, and iii) model refinement. Our proposed diagnostic workflow~(Figure~\ref{figs:workflow}) encompasses the goals of understanding model behavior and diagnosing the model decision space for enabling data scientists and domain experts to generate insights about potential inadequacies in the data and in the model quality. The refinement step is an obvious action as a result of these insights, but is outside the scope of our work.

Analyzing summary statistics of model performance through the lens of visualization techniques is the most common approach for finding matches and mismatches between model predictions and ground truth data. To this end, ModelTracker~\cite{amershi15} provides a unified interface for error detection and debugging for binary classifiers showing item-wise distributions of prediction scores. Bilal \etal propose the confusion wheel visualization~\cite{alsallakh2014visual} and other linked views to show probabilities of items belonging to different classes for multi-class classifiers.  Squares~\cite{ren2017squares} provides a single, unified visualization of performance metrics and easy accessibility to the data for debugging multi-class classifiers. For enhancing the interpretability of classifier predictions, Cortez and Embrechts~\cite{cortez2011opening} use a sensitivity analysis approach for letting users understand the effects of variation of input values on model outputs. While these methods are able to diagnose performance issues 
at the level of a single item~\cite{alsallakh2014visual,amershi15,ren2017squares} or single features~~\cite{cortez2011opening}, they lack a holistic summary of the entire decision space that exposes associations among subsets of items and features, and communicates the reasons behind the model decisions. Through an explanation-based approach, we can let analysts explore these associations for a large, high-dimensional data set, drill-down to individual items, and diagnose potential problems with respect to both global and local decisions. This leads to actionable insights about the limits to which model quality can be improved, and ultimately, hints about how to improve the data.

%% file: tex/motivation.tex
\section{Model Diagnostics}
\label{sec:model-diagnostics}

We use the term \textit{model diagnostics} to indicate the steps necessary for a domain expert or a model developer to semantically validate the decisions made by a model using their domain knowledge. In this section we outline the different goals for a user when using a model diagnostic interface and provide an overview of the implementation of the resulting workflow~(Figure~\ref{figs:workflow}). 
The workflow was derived through a long term collaboration among visual analytic researchers and model developers and domain experts in the medical field, specifically in the application scenario of hospital visits. The over-arching goal in this scenario is to use predictive modeling for reducing patient wait time and optimizing the hospital resources needed for admitted patients.


\subsection{User Goals}

In the course of our interactions with domain and machine learning experts and analyzing a variety of model building problems, we realized that the model diagnostics problem can be decomposed into the following main goals; which we express as a set of questions as shown in Figure~\ref{figs:workflow}.  


\textit{G1: What is the overall accuracy of the model?} In this step, experts need to get an overview of the distribution of prediction scores across the data items, derive an understanding about the uncertainty associated with predictions of certain items, and generally where the predictions are correct or incorrect.


\textit{G2: What are the main decisions the model makes?} A trained classifier creates a decision space that maps a (potentially high-dimensional) input space into the output space defined by the two labels \textit{true} and \textit{false}. Understanding what these decisions are and how frequently they are made is a crucial piece of knowledge domain experts want to draw from the classifier. For instance, in the healthcare scenario we explore in this paper it is crucial for domain experts to know that the vast majority of decisions the classifier makes are based on a small set of drugs (features). They also want to ensure that different sets of drugs are used by the classifier to make decisions about different sets of patients (\eg, a group of patients is characterized by \emph{Ondansetron} and \emph{Sodium Chloride}, whereas another is characterized by antibiotic drugs).

\textit{G3: How accurate are the decisions the model makes?} Together with knowing what decisions the model makes, it is crucial to also know how accurate these decisions are. Using the same example as above, it is not sufficient to know that the model classifies a group of patients according to the drugs they received, but also how often this decisions are correct or incorrect.

\textit{G4: How can one change the data or the model to improve its decisions}? Understanding decisions and assessing their accuracy is relatively useful, but the ultimate goal for a model developer is to actually gain \textit{actionable} insights on how the model can be \textit{improved}. Some of the insights experts want to derive include: whether the model parameters should be tuned or a better set of features should be derived.

In this work we do not provide specific support for the actual parameter tuning or data processing steps necessary to improve the model.
The black-box nature of our approach is illustrated in Figure~\ref{figs:workflow}, which shows that the model diagnostic workflow is an extension of (and not a part of) the existing model building workflows that data scientists follow as part of their routine. Modelers have specific ways and tools to perform these steps and intervening on their established practices is out of the scope of this work. Rather, in this work we focus on providing support for the diagnostic part experts may want to execute at the end of each modeling round and which is currently not well supported by existing tools and practices.
The diagnostic insights produced by our workflow provides hints about whether the input data or the model structure needs to be changed for improving the prediction quality.

\subsection{Workflow}
\label{sec:workflow}


The workflow we propose results from two pre-processing operations: \textit{explanation generation} and \textit{visual mapping}.

\textit{Explanation generation} takes as an input a data set and a trained binary classifier and creates for each instance in the data set an \textit{explanation}. An explanation is a description of the logic (or rule) the classifier uses to assign a given label to the instance. For this purpose, we leverage a method developed by Martens and Provost~\cite{martens2007comprehensible}, which computes, for a given instance which features need to be ``removed" in order to change the classification outcome. For instance, in a text classification problem, an explanation for a document consists of the words that need to be removed in order to change the label originally assigned by the classifier. In Section~\ref{sec:algo} we describe in more detail how the explanation method works.

\textit{Visual mapping} takes as an input the data set and the set of explanations, and builds a set of interactive visualizations~(Figure~\ref{figs:workflow}) that support the user goals we outlined above. The interactive workflow revolves around three main linked interfaces; each one supporting the analysis of model decisions at different levels of granularity and addressing the user goals. 


\par \noindent \textbf{Outcome-level.} The first step focuses on overall accuracy of the model, using a representation similar to a confusion matrix. The main goal of this step is to get a sense of how data distributes across the prediction score computed by the classifier (typically a score between $[0, 1]$), and the four possible outcomes: true or false positive and true or false negative. By visualizing how data distributes across the four possible outcomes the user can gain a sense of how accurate the model is~(\textbf{G1}) and whether errors cluster around particular sets of scores.
    
    
\par \noindent \textbf{Feature-level.} The second step uses the computed explanations to generate an overview of decisions made by the classifier and their accuracy. Each explanation is described by the set of features it uses to explain an instance and, as such, it provides a description of how the model makes its decisions. In this step, we group together all the explanations (and thus the instances) that contain the same set of features, compute accuracy statistics on top of them, and use these groups as a visual interactive summary of the decisions the model makes. By visualizing the explanations and their accuracy the user can get a sense of what are the major decisions the model makes and how accurate they are~(\textbf{G2, G3}).
    
    
\par \noindent \textbf{Instance-level.} The third step focuses on the analysis of a single user-selected explanation and the instances it explains. Once an interesting explanation has been found in the previous step, it is often useful and necessary to drill-down to the individual instances to observe how the data items contained in an explanation distributes in the original data space. Being able to observe their actual data values and the decisions the model enables experts in formulating hypotheses about why the classifier fails to make correct decisions with some instances. In other words, when it is possible to visually compare the data values of instances that have the same explanation but different outcomes, users can draw inferences on the root cause of the diverging outcomes. Therefore, by visualizing single instances the user can reason on how the model makes decisions and derive potentially useful hypotheses about how they can be improved~(\textbf{G4}). 
    

These three steps are linked in a sequence by user-driven filtering mechanisms. The user can select specific sets of values at the outcome-level and visualize them at the features-level. While observing the main set of decisions at the feature-level, she or he can select specific explanations and inspect individual instances in the instance-level interface.


It is important to stress the key role explanations play in the workflow. By computing the explanations and computing statistics on top of them we can effectively provide a description of the main set of decisions the model makes \textit{without} having access to the internal logic of the model. The relevant aspect of explanations is that they compute a compact description of which features the model uses to make \textit{local} decisions for a \textit{subset} of instances. 
For example, in the medical data analysis explored in this work, where each patient is described by the medications he or she received (features) and the classifier predicts whether the patient will be admitted or not, an explanation can identify a group of patients characterized by a small set of medications; that is, the medications the classifier uses to make its prediction.

%% file: tex/algo.tex
\section{Explanation Method}
\label{sec:algo}

\aritranew{Using explanations, we intend to group data items from the perspective of the machine learning model being analyzed.}
In order to do so without relying on a particular model, that is, treating the model as black box, we can estimate which features were involved in its decision making process.
\aritranew{In our initial approach, we had explored alternative methods for grouping the data by looking only at prediction scores of the model~\cite{class_signatures}. However, we realized that those methods mostly reflect the intrinsic structures of the data set instead of the decision making process of the model. Therefore, in this work we build explanations by finding the minimal amount of change necessary to change the prediction of the analyzed model, specifically, a binary classifier. Also, contrary to our previous approach of using explanations to detect only the commonly used features by a model~\cite{rivelo}, here we focus on explanations as a way for experts to diagnose correct or problematic model behavior and address the goals \emph{G1, G2, and G3}, that were outlined in Section~\ref{sec:model-diagnostics}}.

\aritranew{Explanations are created using a} trained model by creating synthetic input values derived from observed data items revealing this input-output relationship.
The set of changes to the values that swayed the outcome of the prediction is then called explanation $e$ for the given original data item:
\[
\min_{e} \, L(v - e) \neq L(v)
\]
where $L$ is the label function with ``positive" or ``negative" as result and $v$ is the data item to be explained.
In order to compute $e$, the prediction function $P$ of the classifier \aritra{is used} with a threshold $t$:
\[
L(v) = P(v) > t
\]
The output of $P$, the prediction score, is a number between 0 and 1 indicating the confidence of a classifier in the predicted outcome.
The threshold $t$ is chosen to yield the most correctly predicted items on the training data.

Prospector~\cite{prospector} and LIME~\cite{DBLP:journals/corr/RibeiroSG16} both propose algorithms that can be used to create explanations. The metric used for minimizing $e$ depends on the explanation technique. Prospector assumes feature independence and thus minimizes $e$ by combining one-dimensional impactful changes of the prediction score. LIME on the other hand creates a local new simpler model by sampling the neighborhood of the analyzed data item and extracts $e$ from this transformed local space. Those two methods aim to approximate minimal explanations in real valued high-dimensional input data spaces.

In our case we are dealing with high-dimensional \textit{binary} input data.
For most applications, like text analysis or movie recommendation, binary input data is sparse, \ie, almost all feature values are 0 instead of 1. Therefore, binary data can also be interpreted as a bag of features. That is, a data item can be treated as set of features whose value is 1.

Martens and Provost~\cite{Martens:2014:EDD:2600518.2600523} provide an algorithm for computing minimal explanations for binary input data.
As Prospector and LIME only generate approximate explanations for data items we adopt and extend this method instead.
The method allows for only removing features from the bag of features.
This restriction comes from the observation that allowing additions to the bag of features can ``tone out" the original item by adding unrelated features with high impact on the prediction score.


The algorithm to generate explanations using this method consists of successively removing features from the bag of features until the prediction outcome changes. The order of the removal is determined by the largest change in prediction score when removing a feature. \textit{The set of features that are removed from the bag of features in order to change the outcome of the prediction is then called an explanation of the original data item.}

One problem with the original algorithm is that it contains a series of conditions that make it give up on explaining some of the \joschi{instances} when these conditions are met. In our case however we want to be able to provide a full picture of the data set and as a consequence we want to create an explanation for every \joschi{instance} provided in the data.
For this purpose we decided to introduce a few modifications to the original algorithm:
%
%
\begin{itemize}
    \item The original algorithm enforces a maximum length of explanations and declares an item as unexplainable if it fails to find an explanation that is shorter than the limit. In our implementation we removed this restriction. The main consequence of this modification is that sometimes the algorithm may produce explanations that are very long and unintelligible. Those explanations however are interesting because they can help us detect and visualize edge cases which may reveal surprising information. In addition, having many long explanations that explain only a few data items can be an indicator \joschi{of a highly complex model with few similar instances or} that the model is overfitting as it is trying to memorize \joschi{individual} labels. 
    
    \item The algorithm can run into plateaus where removing any feature does not change the prediction score. The original algorithm gives up in this case. We circumvent this problem using the following two-step strategy: in this case we select a feature at random and let the algorithm work as usual. Once an explanation has been computed, we follow-up with a ``clean-up" step \joschi{removing} features that do not contribute to a change of the prediction in the end. This extra-step can be very computationally expensive if the input data is not sparse, however, it is necessary to ensure that the resulting explanation is minimal.
    
    \item The original algorithm skips data items whose prediction outcome never changes. As we use explanations as estimate of which features were involved in the decision making process of the model we assign the explanation of those cases to be the original data item. That is, all features present in the original item make up the explanation as \joschi{all} of them were necessary for the model to compute the predicted label.
\end{itemize}
%
The explanation algorithm can take several hours to compute even for small data sets depending on the sparseness of the data.
This requires the generation to be performed offline before analyzing a model.
In order to shorten the computation time we utilized caching of partial explanation results in order to reduce the number of queries to the machine learning model.

%% file: tex/ui.tex
\input{fig/overview}

\input{fig/expl_main}

\section{Visual Interface}
\label{sec:ui}
Our proposed user \joschi{interface}\footnote{\href{https://github.com/nyuvis/explanation_explorer}{https://github.com/nyuvis/explanation\_explorer}} consists of three different panels, each corresponding to the different goals of our proposed workflow that we described in Section~\ref{sec:model-diagnostics}. By interacting with each panel and navigating across these panels, experts can diagnose different aspects of model behavior.

In the visualizations that are a part of our interface, 
the colors orange and blue are used to show negative and positive \textit{prediction} quantities.
A hatching pattern is used for quantities where those predictions are \textit{incorrect} according to the ground truth of the data.
In this section, we describe each panel according to the order of the workflow: \tabA of the machine learning model, the \tabB, and the \tabC.

\subsection{Statistical Summary View}

The purpose of this panel~(Figure~\ref{figs:overview}) is to address \textbf{G1} by providing a quick summary of the performance of a trained model that can help detect shortcomings before proceeding with further analyses of the model. 
The view consists of multiple components.

The histograms~(Figure~\ref{figs:overview}A) show the distribution of data items over prediction scores.
The chosen threshold is shown as vertical line.
Bars going up indicate the number of predicted positive labels while bars going down show predicted negative labels as emphasized by the color of the bars.
The prediction score goes from $1$ to $0$ from left to right to match the order of cells in the confusion matrix.
Likewise, bars at the bottom, left of the threshold, and at the top, right of the threshold, depict incorrectly predicted data items as indicated by their hatching pattern.
Selecting a particular bar lets the user navigate to the \tabB for inspecting items that fall in the given range of prediction scores.

The confusion matrix~(Figure~\ref{figs:overview}B) splits data items by their ground truth (vertical) and the predicted label (horizontal).
The edge of the matrix shows the sums of its columns and rows.
The predicted label depends on a threshold that divides prediction scores into positive and negative.
We choose the threshold to minimize incorrect predictions (\ie, the threshold with the smallest number of false positive and false negative predictions).

The ROC curve on the testing data~(Figure~\ref{figs:overview}C)
shows the false positive rate ($\frac{FP}{FP + TN}$) plotted against the true positive rate ($\frac{TP}{TP + FN}$).
The thresholds for those values are implicit in the plot.
However, the position for the chosen optimal threshold (as described above) is indicated in the plot via two crossing lines.

The area under the ROC curve (AUC) is also shown for both the testing and the training data set.
An AUC of $1$ indicates optimal prediction while an AUC of $0.5$ \joschi{equals} classification by flipping a coin.
Comparing the training AUC to the test AUC is a good estimator of how well the given model generalizes the training data.
A very high training AUC with a much lower test AUC indicates overfitting of the training data.
In addition to the AUC the accuracy of the model with the chosen threshold is also shown.

\subsection{Explanation Explorer}

The second panel, the \tabB (Figure~\ref{figs:expl_main}),  addresses \textbf{G2} and \textbf{G3} by encoding a list of explanations based on the method we described in Section~\ref{sec:algo}. The explanations are representative of the main model decisions and the associated statistics about explained items provide insight into the accuracy of those decisions.
Each row in the list represents one subgroup of data items explained using a single explanation set.
The rows can be filtered based on different criteria for user exploration which we describe below.
The row on top shows information for the full set of current data items.

The first column of the list shows this explanation (Figure~\ref{figs:expl_main}E).
In order to make this information quickly readable we only show the \aritra{first three features of an explanation and indicate if there are more features present by adding a marker, showing the number of remaining features, on the right side of the feature names.}
Furthermore, the feature descriptions are abbreviated in a way that each feature takes up the same amount of space.
With this the complexity of an explanation~\ie, the number of features used in an explanation, can be seen at a glance.
The full description of all features can be seen in the tooltip when hovering over the features.
The design decision to show only up to three features stems from the fact that only short explanations can be easily interpreted and having many long explanations is usually a sign of problems with the classifier, like overfitting, and in that case, the actual features involved are less interesting.

The next column shows the relative distribution of predicted labels of the explained subset of data items as stacked bars (Figure~\ref{figs:expl_main}F).
\joschi{The} colors blue and orange are used to indicate a positive and negative prediction respectively while a hatching pattern indicates incorrect predictions.
The actual numbers are shown in the bars as well.

The bars in the third column show the size of the subset relative to the largest explanation subset of the current data items (Figure~\ref{figs:expl_main}G).
The bars are split according to the distribution of the ground truth labels.
Two shades of gray are used to avoid confusion with distributions of predicted labels.
The total number of items in the subset along with the number of positive items according to the ground truth is written in the column as well.

The fourth column shows the odds ratio of the subset on a logarithmic scale (Figure~\ref{figs:expl_main}H).
Whiskers indicate its confidence interval.
Odds ratio is a popular metric for determining effectiveness in evidence based medicine and clinical trials.
It is computed by comparing the subset explained by the given explanation with the full set of current data items.
\joschi{This way we can detect whether an explanation describes a consistent subset of instances or if the subset appears like a random sample.}
With this the odds ratio is:
\[
\frac{p_e / n_e}{p_t / n_t}
\]
where $p_e$ and $n_e$ is the ratio of positive and negative items respectively in the explanation subset and $p_t$ and $n_t$ is the ratio of those items in the remaining data set.
The confidence interval of the odds ratio is then computed as:
\[
\exp{\left(\log{(\text{odds ratio})} \; \pm \; 1.96 \sqrt{P_e^{-1} + N_e^{-1} + P_t^{-1} + N_t^{-1}}\right)}
\]
where $P$ and $N$ are the actual number of positive and negative items in the explanation subset $e$ and the remaining data $t$.

An odds ratio larger than one indicates that the explained subset is significantly positive with respect to the rest of the current data items.
Likewise, a value smaller than one indicates that it is significantly negative.
However, if the confidence interval crosses one the subset is not significantly different.
To highlight this important special case the odds ratio and the whiskers are drawn in red in this case.

At the right end of each row is a button (Figure~\ref{figs:expl_main}I) to inspect the explained subset more closely in the \tabC as described in Section~\ref{sec:item_level}.

The rows shown by the \tabB can be reordered as well as filtered.
As shown in Figure~\ref{figs:expl_main}, the panel features various controls on the left hand side to accomplish those operations .
Filtering works by first selecting affected rows (either by clicking on a row or by using widgets on the left) and then clicking on the ``+" in the list of filtered data items~(Figure~\ref{figs:expl_main}B)
Each new filtering of data items creates a new entry in this list showing the current number of data items.
By selecting entries higher up in the list the user can go back to this filter.
The topmost entry always contains all data items of the entire data set.

Besides getting a filter for a given prediction score range from the \tabA there are two ways of filtering items: searching and conditioning.
The search field~(Figure~\ref{figs:expl_main}A) can be used to select rows whose explanation matches the query specified by the user.
While typing or using arrow keys suggestions for feature names are shown in a dropdown list.
Those suggestions are sorted by how often that feature appears in explanations and how closely it matches the already specified query.
The query can contain multiple features that need to appear in the explanation separated by a comma ``,".
The conditioning widget~(Figure~\ref{figs:expl_main}C) allows to filter by quantities.


As shown in Figure~\ref{figs:expl_main}D, different metrics can be used to filter or reorder the list of explanations.  
A good use for the conditioning filter is to remove explanations that only explain a small subset of the data when looking for unusual or significant subsets.
The explanation rows can also be reordered using these metrics. 
The widget contains a list that shows the order in which explanations get sorted.
Each element has a symbol next to it indicating the sort direction which can be clicked on to change the sort direction.
Selecting an element brings it to the top of the list.

The metrics used for reordering are the same as those used for conditioning with the additional option of lexicographical sorting by using the feature names of the explanations.
Common metrics to use for sorting or reordering, besides total amount of items, are ``uncertainty" and ``odds ratio".
``Uncertainty" (the closeness of the odds ratio to one: $-| \log{(OR)} |$) provides a view into problematic areas of the machine learning model sometimes even unpredictable items when items with the same value configuration have different ground truth labels.
``Odds ratio", based on the computation mentioned earlier, points to especially strong predictive areas of the machine learning model.

\subsection{Item Level Inspector}
\label{sec:item_level}
The third panel~(Figure~\ref{figs:inspect_all}) allows for a more granular inspection of items explained by a given explanation set. This addresses \textbf{G4} by providing hints about the extent to which a model can be improved and if changing the data is necessary for that purpose.
The panel consists of a matrix showing the actual feature vectors of the given items.
Each row represents a unique feature vector pattern while columns represent features.
Rows can be expanded so that each row represents exactly one item.
Cells in the matrix are filled if the corresponding feature vector contains the feature represented by the column.
As rows are aggregates of multiple data items, the number of items is shown as a bar with the number indicated on the left side of the matrix.
The feature names for the columns are shown slanted on top of the matrix.
Bars behind the names show how often the feature is present.
The very first column in the matrix shows the predicted label (using the colors blue and orange) and its correctness (hatching pattern for incorrect predictions) of the given data item.
Items with the same feature vector configuration but different labels are shown in different rows.

Rows and columns can be reordered using different options, similar to \tabB.
One of the important reordering criteria for rows is the \textbf{feature order}, where items are ordered by seeing whether the first feature of the columns is present and then if the next feature is present, and is repeated for all the columns.
An important reordering criteria for the columns is the \textbf{relative feature importance}: the gini feature importance with respect to the current subset of data items and their predicted labels and correctness.



The combination of the ``feature order" and the ``relative feature importance" criteria provide a particularly interesting view on the subset of data items.
Using this order, the most discriminating features with respect to predicted and actual labels are shown first.
\joschi{Since} the rows are ordered by those features, a user can follow those orderings to see how to separate different predicted and actual labels.
This guidance of the user to relevant associations in the item subset are useful for quickly understanding the raw data.
Note that it is sometimes possible to fully separate data items this way.
However, utilizing this separation would be overfitting on the validation set.
Furthermore, the opposite situation with exact same feature vectors but with different labels that cannot be separated exists as well.

Some features are not discriminative in terms of ``relative feature importance". They can be ignored to simplify the matrix view.

%% file: fig/overview.tex
\begin{figure}[t]
\centering
\includegraphics[width=\columnwidth]{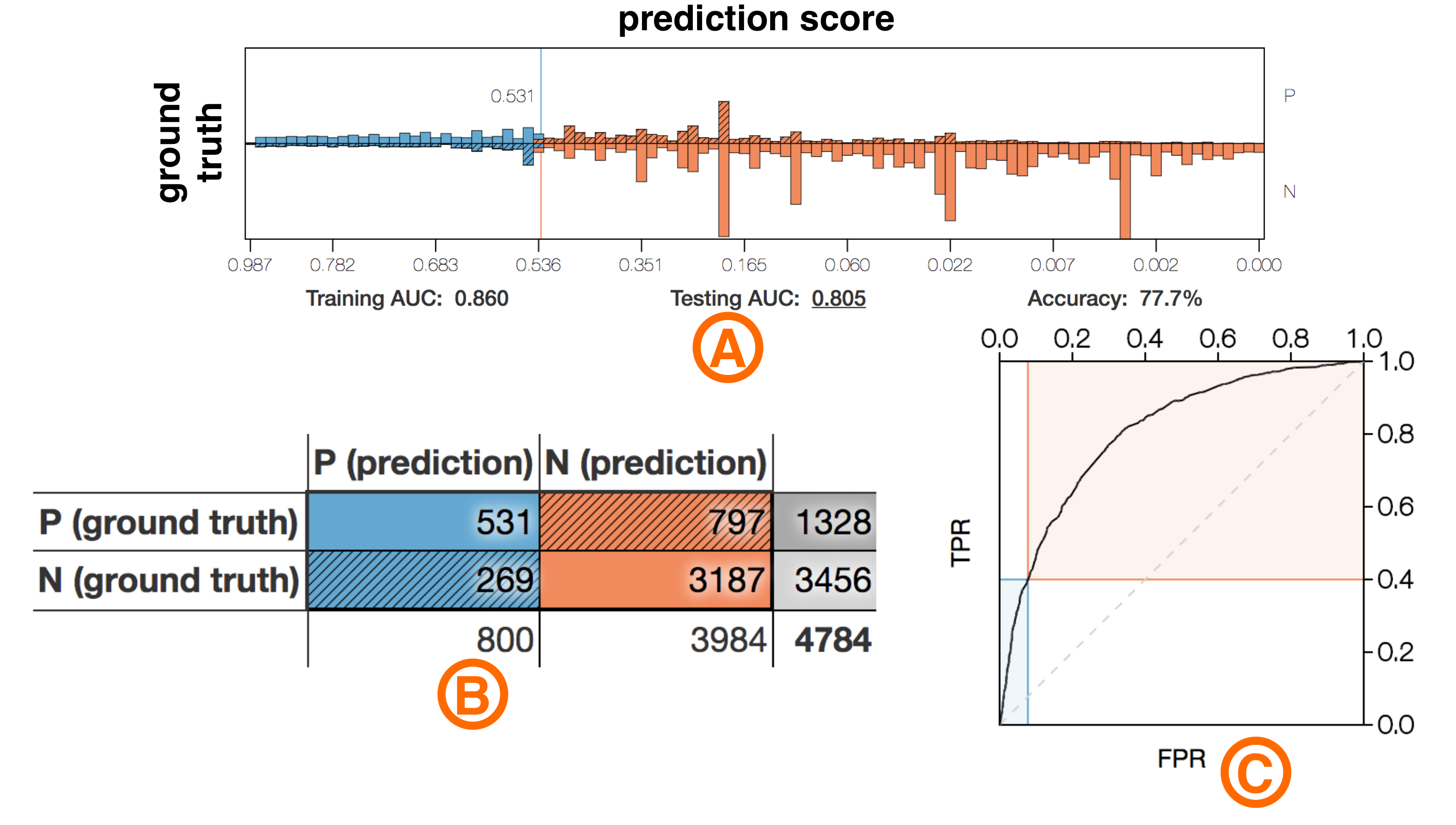}
\vspace{-8mm}
\caption{
The \textbf{\tabA}.
(A) Histograms showing the distribution of prediction scores.
The direction of the bars indicates the ground truth and their position relative to the threshold line (at 0.531) indicates the predicted label.
(B) The confusion matrix shows the number of correct and incorrect predictions. (C) The ROC curve shows the prediction quality.
}
\vspace{-5mm}
\label{figs:overview}
\end{figure}


%% file: fig/expl_main.tex
\begin{figure*}[t]
\centering
\includegraphics[width=0.8\linewidth]{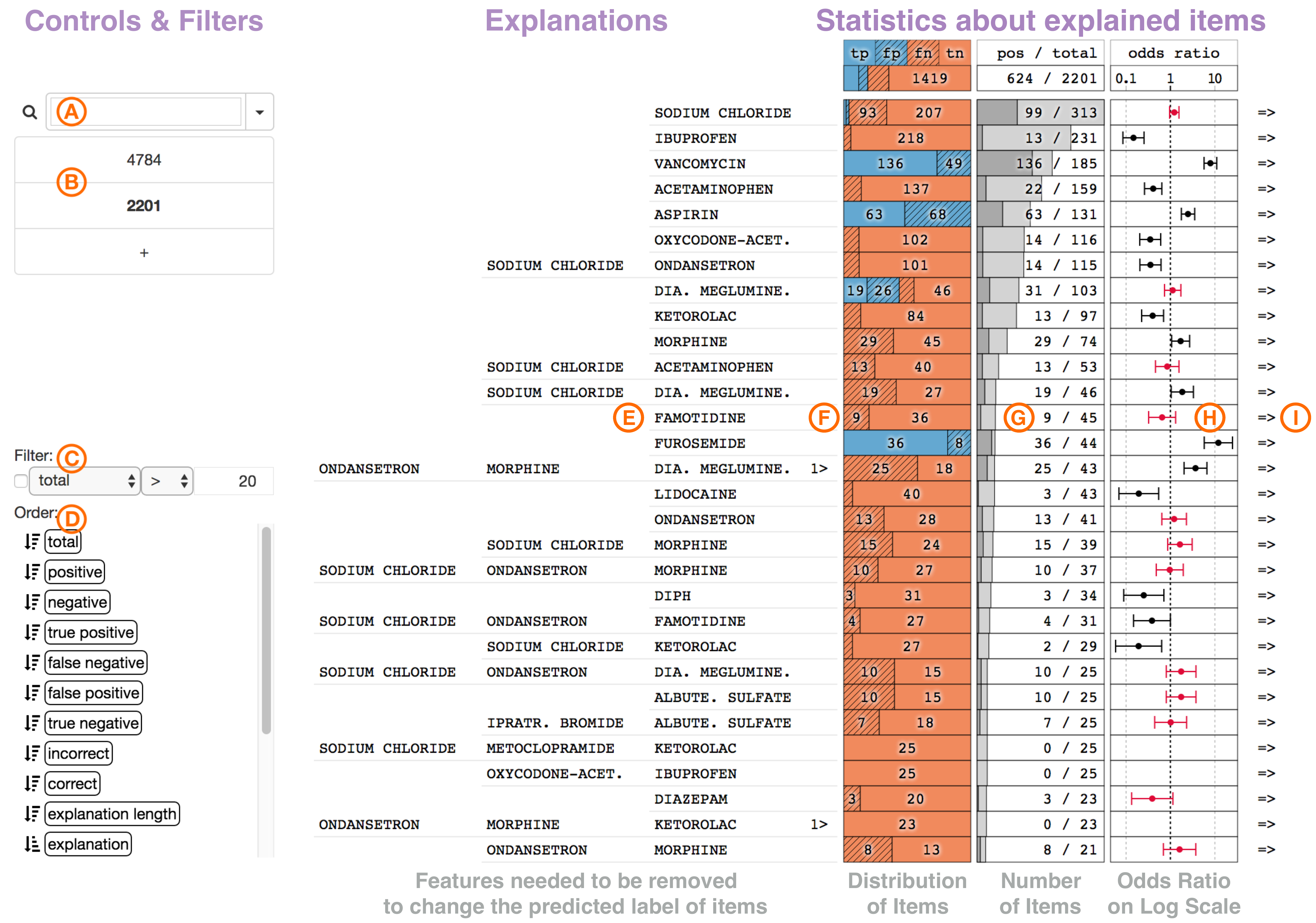}%
\raisebox{2.05em}{\includegraphics[height=6.9em]{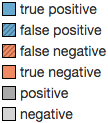}}
\vspace{-3mm}
\caption{
In the \textbf{\tabB} each row represents a group of data items explained by a set of features (E).
An indicator is shown for explanations longer than 3 features.
Column (F) shows the distribution of true / false positive / negative data items within the group.
Colors show the predicted label (``blue" for positive and ``orange" for negative) and a hatching pattern indicates incorrect predictions.
Column (G) shows the number of items captured by the explanation.
The bars are relative to the size of the largest explanation.
Column (H) shows the odds ratio of the group on a logarithmic scale.
Whiskers show the confidence interval.
The arrows on the right (I) navigate to the \tabC focusing on the given explanation. The controls of the \tabB are shown on the left.
The first entry of the list of filtered data items (B) represents the full dataset and following entries show sizes after filter steps are applied.
The ``+" creates a new filter according to the current selection of explanations.
Explanations can be selected satisfying a condition (C) or by searching for features in the search box (A).
The sort order of explanations is defined by the list at the bottom (D).
}
\vspace{-5mm}
\label{figs:expl_main}
\end{figure*}

%% file: tex/case.tex
\section{\aritra{Case Study: Expert Analysis of Medical Outcome}}
\label{sec:case_study}

The proposed workflow described above stem from a one year collaboration with a machine learning expert and a medical doctor from the \textit{NYU Langone Medical Center}, \joschi{both co-authors of this paper}. The medical machine learning team at the medical center works in tight collaboration with doctors and hospital management to derive novel methods to automate medical procedures, provide diagnostics support, improve efficiency and gain novel insights on medical procedures and processes.

\noindent \textbf{Domain Problem Description.}
Our collaboration focused on the analysis and improvement of models built to optimize processing times in the emergency department of the hospital. The crucial decision here is whether a patient coming to the emergency room will end up being admitted to the hospital or sent home. In the case of a patient being admitted to the hospital, a bed has to be prepared for the patient which results in a 2-hour waiting period where the patient occupies a bed in the emergency room preventing other patients \joschi{from being} processed. If the waiting time can be reduced by knowing early if a given patient will be admitted, the throughput of the emergency room can be increased.

\input{fig/inspect}

The idea to reduce this wait time is to use predictive modeling at the earliest time possible so that an admitted patient can be moved sooner. The amount of data available to make this decision however is very limited. When first presented with a patient the emergency doctor orders medication for treating, stabilizing, or preparing the patient for procedures or tests and eventually will conclude a diagnosis and decide whether the patient is in need of admission.

As medication is the earliest recorded indicator of the admission result and also is recorded before lengthy procedures or tests it is the most promising candidate for a predictive model. The main machine learning task is therefore to verify whether a viable model can be built by using exclusively the limited information available.

Other work has been done in this regard, however, using input features that are not readily available (\eg, information from medical notes that are written after the fact) or are hospital specific (\eg, mode of arrival, triage score) \cite{pmid21705374,pmid24421344,pmid24509606}.


\input{fig/odds_ratio}

During our collaboration the team of visualization experts met with the medical team regularly to understand the problem and the data, and to develop collaboratively visual analytics solutions for model diagnostics and interpretation. The workflow we described in the paper resulted from numerous iterations over the methods used to derive information from the model and the methods used to enable their interactive visual exploration.

In this section, we describe one particular example that showcases the capabilities of the proposed method and provides insights on how it is able to support diagnostic analysis of complex machine learning model used in a relevant real-world scenario. \aritra{In the following, the term ``we" is used to refer to the team of visual analytic experts, a machine learning expert, and a medical doctor, who collaboratively worked on the usage scenarios described below.}

\noindent \textbf{Selecting Initial Data and Model (G1).}
We initially gathered a dataset of 5980 patients ($28\%$ admitted) with binary vectors indicating medications given to the patient. Those patients were randomly split into a training (1196 patients with $30\%$ admitted) and test (4784 patients with $27\%$ admitted) dataset. We then computed several models and tweaked them using mostly the \tabA and model specific approaches.
\joschi{This initial dataset contains 398 unique medications.
The table below shows a summary of the models we trained and their performance.}






\begin{center}
\begin{tabular}{l|rr}
Model & Training & Test AUC \\
\hline
Gaussian Na\"ive Bayes (GNB) \cite{DBLP:conf/flairs/Zhang04} & 0.58 & 0.52 \\
Logistic Regression (LR) \cite{Yu:2011:DCD:2039082.2039098} & 0.85 & 0.79 \\
Random Forest (RF) \cite{Breiman:2001:RF:570181.570182} & 0.88 & 0.79 \\
Multi Layer Perceptron (MLP) \cite{DBLP:journals/corr/HeZR015} & 0.85 & \textbf{0.80} \\
\end{tabular}
\label{tab:auc1}
\end{center}

As we can see most of the models achieve similar performance on the test data. In the following we focus exclusively on the \textit{Multi Layer Perceptron} model but the same kind of analysis can be performed on any of the other models \joschi{with similar results for the models with similar predictive power}.


\noindent \textbf{Exploring model decisions and spotting problems (G2 \& G3).} 
To start the analysis we compute all the explanations and visualize them in the \tabB shown in Figure~\ref{figs:expl_10_size}, which by default is sorted by frequency of explanations. The first thing we notice is that \emph{Sodium Chloride} is the most common explanation and that it contains a considerable number of misclassified instances.

\emph{Sodium Chloride} represents an intravenous therapy, the infusion of a liquid directly into a vein. As part of a medication order it is used to increase the effectiveness and response time of a drug and also to apply medication if a patient is unconscious. Used by itself it has the only purpose of hydrating a patient.

\joschi{The distribution for the explanation shows both positive and negative predicted outcomes}\aritra{, which may seem paradoxical at first.} \enrico{This result however stems from the fact that the context of an explanation (that is, whether features co-occur with the features used in the explanation; note that certain co-occurring features form other explanations as they have a direct influence on the outcome) matters in terms of which outcome it explains.} 

\enrico{The \tabC can help us clarify this situation.} We can see that hospital admission is the predicted outcome when \emph{Sodium Chloride} appears together with other drugs, whereas when this is the only medication the patient received, the patient is predicted to get sent home (Figure~\ref{figs:inspect_all}b).


\joschi{Looking} at the odds ratio value for this explanation we also notice that this subset is not significantly predictive and that the misclassification rate is high (weak signal). Note that even though \emph{Sodium Chloride} is the most common explanation it cannot be used as a significant indicator of the outcome. \joschi{From a medical perspective this makes sense as \emph{Sodium Chloride} is mostly used as supporting medication, however, the machine learning model still assigned predictive power to it. This indicates that the data did not contain a strong enough signal to make a more informed decision in those cases.}

Another common explanation is \emph{Ibuprofen} a pain relieving drug.
It is predictive for non-admissions which is likely due to patients with pain symptoms that turned out to be benign.
The odds ratio indicates a significant relation to the outcome.
On the other hand \emph{Vancomycin}, an antibiotic used for treating infections, is significantly linked to hospital admission which is expected.

After filtering out uncommon explanations ($< 20$ explained items) ordering the explanations by ``odds ratio" reveals significant indicators for admission and non-admission (Figure~\ref{figs:expl_10_or_pos}).
In addition to the already discovered significant explanations we can see \emph{Furosemide}, a drug for treating congestive heart failure, as being strongly indicative for admission and certain drugs in combination with \emph{Sodium Chloride} strongly linked to non-admission (Figure~\ref{figs:expl_10_or_neg}).
The drugs in question are pain-relievers (\emph{Morphine} and \emph{Ketorolac}) and drugs to help with stomach problems (\emph{Ondansetron} and \emph{Metoclopramide}).
Note that using an IV \joschi{(\emph{Sodium Chloride})} for stomach related problems helps both hydrate the patient and ensures the intake of the medication (after \eg, vomiting).

\par \noindent \textbf{Finding Weaknesses (G4).}
Ordering explanations by ``uncertainty" (Figure~\ref{figs:expl_10_uncertain}) shows explanations whose predictions are not significant.
This is often the case when it is impossible to correctly predict a set of identical \joschi{instances} that have \joschi{a contradicting ground truth}.

The first two explanations \emph{Ipratropium Bromide}, \emph{Albuterol Sulfate} (medication for treating chronic obstructive pulmonary disease and asthma, lung diseases that can have chronic and acute symptoms the latter of which requires immediate attention) and \emph{Sodium Chloride}, \emph{Ondansetron}, \emph{Morphine} are both predicted negative.
However, the ground truth of those subset has the same distribution as the overall dataset (thus an odds ratio close to 1).
This means the true admission rate of those two subsets is independent of the medication in question as the admission rate matches the admission rate of the dataset.
If more patients would be observed in the data this rate would likely stay the same.
Through \tabC we can see that the features of the explanations are the only features in the respective data items.
No further information is provided that could help swaying those subsets in a definite direction of admission or non-admission.


Another problematic drug is \emph{Diatrizoate Meglumine} which has a high misclassification rate and an odds ratio close to 1.
The drug is a contrast medium that is given in preparation of PET (positron emission tomography) or CT (computerized tomography) scans.
As the outcome of the scan is not known it cannot be determined whether the test was positive for the hypothesis made by the attending physician.
Furthermore, even the presence of other drugs is no indicator for admission as it only shows the doctor's risk assessment \textit{before} the test was ordered and therefore does not include whether the doctor's assumption was correct.
Note, that Figure~\ref{figs:inspect_all}a shows how outcomes can be better separated using available features.
However, doing so would result in overfitting on the validation data set which should be avoided in any case.

Faced with this revelation we explored how we could provide more information to reduce those ambiguities.
In order to properly deal with cases like \emph{Ipratropium Bromide} and \emph{Albuterol Sulfate} or \joschi{\emph{Sodium Chloride} and} \emph{Diatrizoate Meglumine} more information is needed.
\joschi{Through domain expertise we can reason about the underlying shortcomings of the current dataset, \eg, the nature of the limitations of \emph{Diatrizoate Meglumin}.
In order to overcome those limitations we need to include additional information in our dataset.
For example, including information about the final diagnosis of a patient resolves the ambiguities of patients explained by \emph{Diatrizoate Meglumin} and other problematic explanations mentioned above, and likely improves the overall quality of the prediction\footnote{I\joschi{ncluding other information, such as, mode of arrival, gender, or age, might improve accuracy but would not solve the issues mentioned above.}}.
However, this also} moves the time of the prediction closer to the point in time when the actual decision, whether the patient is admitted to the hospital, is made thus reducing the time-gain for preparing a bed in case of admission.

\par \noindent \textbf{Changing Data and Model.}
\joschi{In the following we describe how we could improve the prediction task by including additional information to our dataset. This additional information, \ie final diagnoses, was added to overcome limitations posed by medications not strongly linked to an outcome, as described above.}
\joschi{In order to include those diagnosis} features in the data we had to capture new data which also allowed for capturing a bigger dataset.
The new dataset contains 154580 patients  ($20\%$ admitted) and was randomly split into a training (30916 patients with $20\%$ admitted) and test (123664 patients with $20\%$ admitted) dataset.
\joschi{It contains 1709 unique medications and 15422 unique diagnoses.}

The best results of different models on the new dataset are:

\begin{center}
\begin{tabular}{l|rr|rr}
\multicolumn{1}{c}{} & \multicolumn{2}{c}{no diagnoses} & \multicolumn{2}{c}{incl. diagnoses}\\
Model & Training & Test AUC & Training & Test AUC\\
\hline
GNB & 0.51 & 0.49 & 0.75 & 0.66 \\
LR & 0.71 & 0.67 & 0.93 & 0.88 \\
RF & 0.69 & 0.68 & 0.98 & 0.83 \\
MLP & 0.71 & \textbf{0.68} & 0.95 & \textbf{0.88} \\
\multicolumn{5}{c}{{\scriptsize Maximum values are chosen using digits not shown.}} \\
\end{tabular}
\end{center}

Again we are focusing solely on the Multi Layer Perceptron model for further analyses \joschi{(even though similar results can be found with the other equally} \aritra{well performing models)}.
In order to compare our new data to the previous dataset we first created models that do not utilize the newly added diagnoses.
However, the resulting AUC is much lower than for the initial data.
Looking at the \tabA reveals a strong concentration of data points at a specific prediction score.
Focusing on this prediction score in the \tabB (Figure~\ref{figs:adm_10_size}) shows that it corresponds to the 62776 patients that did not receive any medication at all.
This configuration predicts non-admission as it is more likely to get sent home when not receiving any medication.
The unusual large number of such cases ($\sim\hspace{-0.2em}50\%$), however, hints at a possible capturing error which would also explain the 11394 cases where patients were admitted.
This failure rate severely affects the machine learning models.
For comparison the next largest explanation of \emph{Ibuprofen} in the new dataset consists only of 2011 patients.
In fact patients without medication were not captured in the original dataset and removing them from the new dataset increases the best train / test AUC to $0.83$ / $0.80$ similar to the original dataset.
For further analysis we include patients without medication.
Utilizing diagnoses in the models strongly increase the possible AUC.

\par \noindent \textbf{How Did Diagnoses Features Change the Model?}
The \tabB of the best model utilizing diagnoses features, Multi Layer Perceptron, can be seen in Figure~\ref{figs:adm_10_diag}.
Noticeably, almost all explanations now consist of diagnoses.
This also means that medication features have now become almost irrelevant except for medications, like \emph{Ibuprofen} and \emph{Vancomycin}, that were strong indicators before.
The most significant diagnoses, using odds ratio, for admission are \emph{Sepsis}, \emph{Sepsis due to unidentified organism}, and \emph{Small Bowel Obstruction}.
Diagnoses that require antibiotics (\eg, \emph{Vancomycin}) and pain medication (\eg, \emph{Ibuprofen}) respectively.
Contradictory or insignificant medications, like \emph{Diatrizoate Meglumine} or \emph{Ipratropium Bromide} and \emph{Albuterol Sulfate}, do not show up anymore as they can be more effectively replaced by their diagnoses.
The largest explanation, with 2619 patients, is \emph{Unspecified} which, after some research, turns out to be due to a policy change before which doctors were allowed to omit a diagnosis if the patient got admitted to the hospital.
Why only 2105 ($\sim\hspace{-0.2em}80\%$) were actually admitted to the hospital remains unclear.

\input{fig/adm}

\par \noindent \textbf{Diagnostic Insights.}
By adding diagnoses to the dataset a strong increase in predictive quality was achieved.
However, seeing that diagnoses effectively replace medication in their predictive power suggests that the ``labels are leaking".
That is, since doctors make the decision of whether to admit a patient at the time of the final diagnosis there is a strong correlation between the label and the features.
This is an undesired effect as the model is not predicting the outcome anymore but merely building an approximate lookup table for diagnosis admission rates.
If the model would have kept using medication and only consulted diagnoses for ambiguous cases the usability of the model would have been improved due to diagnoses.
This is not the case.
Despite its lower objective quality the model using only medications as input emerged as the more practically useful model.
\joschi{
Since experts know about the strengths and weaknesses of the model, they can distinguish between confident and ambiguous cases early and decide whether to accept the prediction or wait for the final decision made by the doctor.
This demonstrates that a statistically weaker model can be more useful in practice.}

%% file: fig/inspect.tex
\begin{figure}[b!]
\centering
\includegraphics[width=\linewidth]{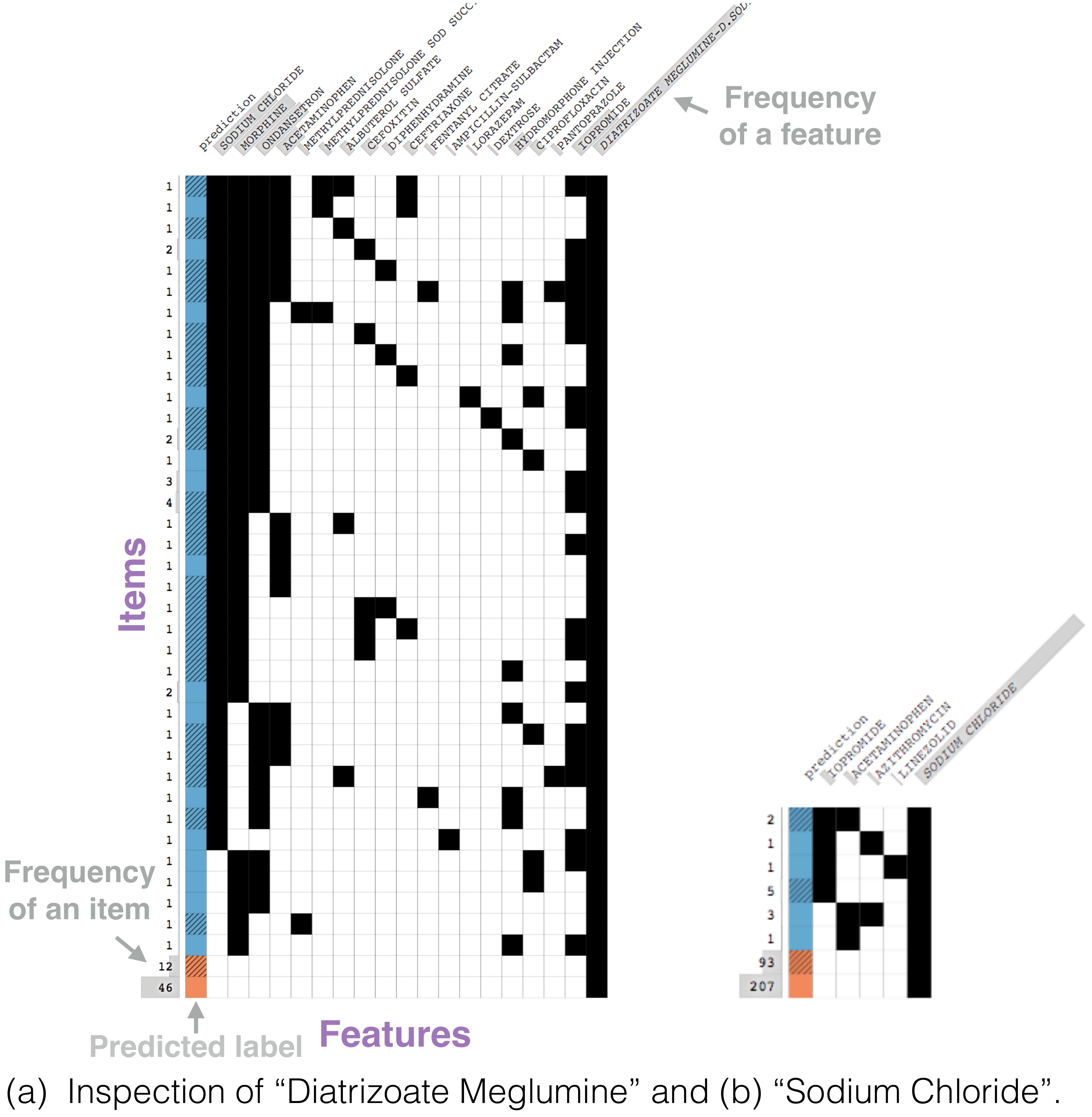}
\vspace{-5mm}
\caption{
The \textbf{\tabC} showing a matrix of data items as rows \joschi{and} features as columns for the explanations \emph{Diatrizoate Meglumine} and \emph{Sodium Chloride} in the initial data set of the case study (Section~\ref{sec:case_study}).
Rows group identical \joschi{instances} together and show the count on the left side.
Features are sorted by ``relative feature importance" showing from left to right how labels can be separated.
}
\label{figs:inspect_all}
\end{figure}

%% file: fig/odds_ratio.tex
\begin{figure*}
\centering
\begin{subfigure}[b]{0.49\linewidth}
    \hfill
    \includegraphics[height=9.5em]{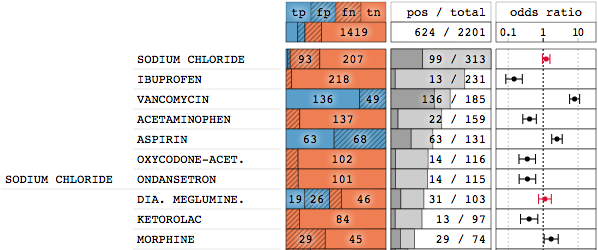}
    \caption{~Ordered by ``total" size showing the most common explanations.}
    \label{figs:expl_10_size}
\end{subfigure}
\hfill
\begin{subfigure}[b]{0.49\linewidth}
    \includegraphics[height=9.5em]{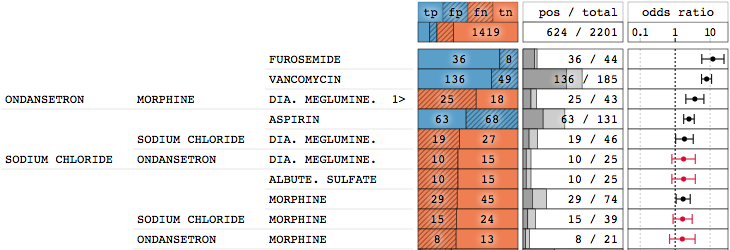}
    \caption{~Ordered by ``odds ratio" showing significantly positive explanations.}
    \label{figs:expl_10_or_pos}
\end{subfigure}%
\\
\begin{subfigure}[b]{0.49\linewidth}
    \hfill
    \includegraphics[height=9.5em]{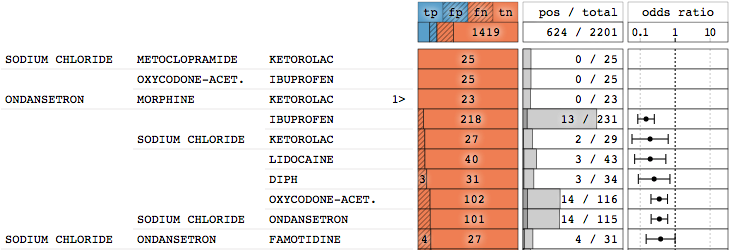}
    \caption{~Ordered by reverse ``odds ratio" showing significantly negative explanations.}
    \label{figs:expl_10_or_neg}
\end{subfigure}%
\hfill
\begin{subfigure}[b]{0.49\linewidth}
    \includegraphics[height=9.5em]{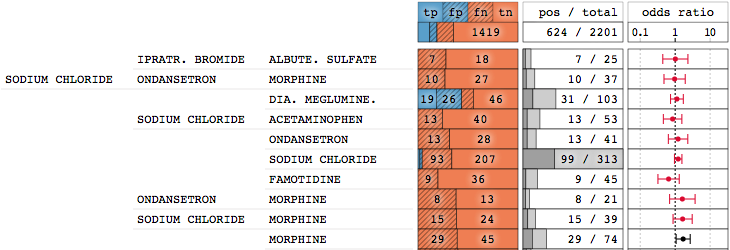}
    \caption{~Ordered by ``uncertainty" showing item subsets whose predictions are not significant.}
    \label{figs:expl_10_uncertain}
\end{subfigure}
\vspace{-3mm}
\caption{
Showing different orders in the \textbf{\tabB} for addressing the goals (G2 \& G3) in the case study (Section~\ref{sec:case_study}). The initial dataset is filtered for explanations with $> 20$ data items.
}
\vspace{-5mm}
\end{figure*}

%% file: fig/adm.tex
\begin{figure*}
\centering
\begin{subfigure}[b]{0.49\linewidth}
    \centering
    \includegraphics[height=9.5em]{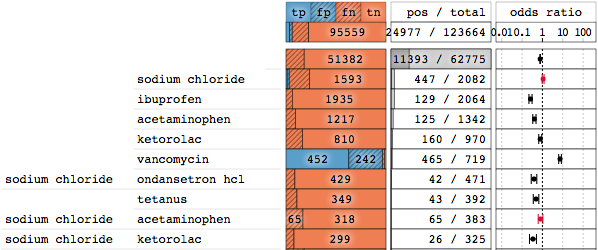}
    \caption{~The second dataset without diagnoses ordered by ``total" size.}
    \label{figs:adm_10_size}
\end{subfigure}
\begin{subfigure}[b]{0.49\linewidth}
    \centering
    \includegraphics[height=9.5em]{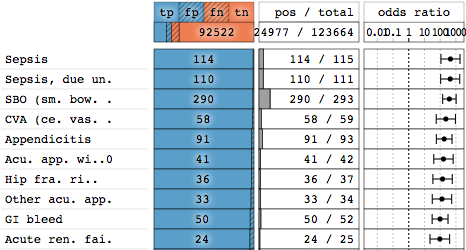}
    \caption{~The second dataset using diagnoses ordered by ``odds ratio".}
    \label{figs:adm_10_diag}
\end{subfigure}%
\vspace{-3mm}
\caption{Showing the second dataset of the case study (Section~\ref{sec:case_study}) with and without using diagnoses features in the \textbf{\tabB}.}
\vspace{-5mm}
\end{figure*}

%% file: tex/conclusion.tex
\section{Discussion}
\label{sec:discussion}









Through our case study of patient visits, we have shown that by aggregating model decisions through explanations, we are able to make sense of a large number of interesting decisions: some expected and some unexpected; some useful and some less useful; and finally some leading to actionable knowledge and some requiring more introspection on the part of domain experts. This level of transparency is necessary for experts and data scientists to built trust in a model and, especially, generate ideas on how it can be improved.

In our interactions we have also noticed the usefulness of using explanations as the main method to make sense of model decisions. As long as the features used for the problem can be interpreted by the user, the concepts expressed in the visualization are easy to grasp and learn. During our collaboration we have experimented with other structures such as trees and rules but we often found that these were either too complicated or hard to use for modeling complex phenomena reliably and succinctly.

As we observed in the case study presented in Section~\ref{sec:case_study} it is important to understand which decisions a model is most certain about and also find the decisions about which it is uncertain. When issues are detected there are several possibilities: training a better model, finding better data, introducing new and more informative features, or deciding that the model can make decisions only for the subset of cases the experts are most certain about. One possible outcome is also deciding that the problem is simply too complex and that expert judgment is, at the current stage, preferable.

From the experience we gained in this project we drew a number of important lessons, which we outline below.

\par \noindent \textbf{\aritra{Lessons Learned.}}
In our work we noticed that many of the issues we spot in our analysis cannot be corrected simply by training a better model with the same data, but need some major redesign of the feature space and a careful analysis of the biases contained in the data. In turn, while diagnosing one or more models built on one data set and set of features can bring useful knowledge, ultimately solutions often have to come from better data engineering. We believe visual analytics can and should play a major role in this regards and find ways to support analysts explore alternative data and feature spaces. This is even more relevant when we observe that visual analytics systems and research tends to focus on one single data set and one single set of features.
Focusing on supporting external changes of data and models offers many challenges and opportunities for visual analytics.






Another important observation pertains to the practical value of developing a visual validation system \joschi{separated from and not interfering} with the existing modeling pipeline. \joschi{From} Figure~\ref{figs:workflow} it may seem natural to envision visual analytics methods able to support the user in closing the loop and apply direct modifications to the model in order to improve it. This is the type of solution advocated by the \textit{interactive machine-learning} paradigm~\cite{amershi2014power}, in which the user can directly instruct the model on how to improve its decisions.

However, through our collaboration, we realized that modelers and experts often have very specific tools they use for model development and refinement and it is often hard to intervene on their familiar processes and infrastructure. A much more viable solution is to develop a methodology that does not require a substantial modification of their existing workflow and infrastructure.

We also observe that while this type of paradigm is useful to provide better examples to the model, it cannot solve the data acquisition shortcomings we have outlined above. Fixing these problems requires domain experts to rethink the whole approach of the stated machine learning problem. For example, improving the input data might require to capture new features from different sources or rethinking of pre-processing steps.
It seems important to figure out in future research which particular settings are the most appropriate for the ``out of the loop" solution we proposed here and which are more amenable to the interactive machine learning paradigm.

A final observation is how the process of validating the model often leads to generating insights that pertain more to the reality being modelled than the model itself. In several occasion, our collaborators ended up spotting potential issues with how their patients are handled in the hospital. Typical examples include situations in which some patients are discharged and at the same time are given medications that represent a strong signal for a serious condition for the doctor.
These kind of mismatches between the mental model of the doctor and the reality modeled is a potential source of process improvement and can be used to take important actions.

In relation to this last observation, it seems interesting to reflect on how visual analytics can \joschi{further} leverage the power of modeling for exploratory data analysis and data sense making. While many 
systems focus on direct visualization of raw data as overview, there seem to be relevant opportunities on using modeling as a preparatory step so that the resulting visualization contains more signals about hidden associations among features and items in the data.

\par \noindent \textbf{\aritra{Limitations.}}
\joschi{The} workflow and \joschi{its implementation} we described work exclusively with sparse binary data and binary classification. Although, explanation generation can be extended to other input data types the visual representation of those explanations has to be redesigned in order to accommodate other data types. Similarly, handling \joschi{classification for} more than two classes is also not trivial.

\joschi{Our} method works only with interpretable features, that is, features have a direct connection to a reality the user can easily understand. Many relevant machine learning problems however require the use of highly non-interpretable features. Classification of images, audio, and video, is a classic example of this case. In these settings the single features used by the model do not have any direct interpretation the user can directly use for model understanding.

\joschi{Our} solution works best with analyzing one single model at a time but it does not provide direct support for \textit{model comparison}. In many practical cases modelers like to train multiple models and then figure out how they compare. While in practice most of these comparisons are currently performed on statistical aggregations, it would be useful to develop methods able to compare multiple models in terms of the \textit{decisions} they make and how they differ. This is even more important in those cases in which models display a similar performance but actually differ in the way the decisions they make.

\joschi{Merging same explanations with different outcomes, like in the case of \emph{Sodium Chloride}, was done to make a user aware of this case. However, merging penalizes the odds ratio. In the cases presented in this paper the odds ratio did not get affected as the correctness for both outcomes were similar. If, for example, the positive prediction were always right but the negative prediction equivalent to a random guess both cases would be underrepresented by the odds ratio.}

\joschi{With respect to scalability, neither the total number of features nor the total number of instances is limiting, since only a subset of available features appear in explanations and many instances are aggregated. However, it can happen that explanations are consistently long or do not aggregate well. This is mostly dependent on the model. Long explanations can be a sign of overfitting or a highly complex model with few similar instances. Explanations in the latter case are less interpretable which demands for a strategy to simplify or shorten explanations.}

\section{\aritra{Conclusion} \& Future Work}

We demonstrated how visual explanations can be effectively leveraged by \aritra{data scientists} and medical experts for diagnosing model decisions and for ultimately making informed judgment about associations among medications and patients' diagnoses. 
We will extend our method to non-binary data and multi-class problems. \aritra{We will also extend our solution for letting data scientists compare explanations from multiple models and leverage our model-agnostic workflow for making informed choices about choosing machine learning models in real-world application scenarios.}



